\documentclass{article}

\usepackage{iclr2027_conference,times}
\iclrfinalcopy  

\usepackage{amsmath,amsfonts,bm}

\def\eqref#1{equation~\ref{#1}}

\def\1{\bm{1}}

\DeclareMathAlphabet{\mathsfit}{\encodingdefault}{\sfdefault}{m}{sl}
\SetMathAlphabet{\mathsfit}{bold}{\encodingdefault}{\sfdefault}{bx}{n}

\DeclareMathOperator*{\argmax}{arg\,max}
\DeclareMathOperator*{\argmin}{arg\,min}

\usepackage{amsmath,amssymb,amsfonts}
\usepackage{amsthm}
\newtheorem{definition}{Definition}
\usepackage{graphicx}
\usepackage{booktabs}
\usepackage{multirow}
\usepackage{url}
\usepackage{listings}
\lstdefinestyle{prompt}{
  basicstyle=\ttfamily\scriptsize,
  breaklines=true,
  breakindent=0pt,
  columns=fullflexible,
  keepspaces=true,
  showstringspaces=false,
  frame=single,
  framesep=4pt,
  xleftmargin=4pt,
  xrightmargin=4pt,
  aboveskip=6pt,
  belowskip=2pt,
  literate={—}{{-}}1 {–}{{-}}1 {“}{{"}}1 {”}{{"}}1 {‘}{{'}}1 {’}{{'}}1 {∈}{{$\in$}}1 {≤}{{$\le$}}1 {≥}{{$\ge$}}1 {×}{{$\times$}}1 {→}{{$\rightarrow$}}1 {·}{{$\cdot$}}1 {…}{{...}}1 {²}{{$^2$}}1
}
\usepackage{hyperref}

\graphicspath{{images/}}

\title{GRADS: Graph-Regulated Agentic Data Synthesis with Controllable Difficulty}

\author{%
Gaocheng Yang\textsuperscript{1,\,*,\,\dag} \quad
Ruijie Tao\textsuperscript{1,\,*} \quad
Kai Ye\textsuperscript{2} \quad
Senhua Zhang\textsuperscript{3} \quad
Zhengqin Zhu\textsuperscript{3} \quad
Ziyang Li\textsuperscript{1,\,\ddag} \\[0.4em]
\textsuperscript{1}Johns Hopkins University \quad
\textsuperscript{2}Nanjing University \quad
\textsuperscript{3}Amazon \\
\texttt{\{gyang41,rtao12\}@jh.edu} \quad
\texttt{kaiye@smail.nju.edu.cn} \\
\texttt{\{senhuazh,tozhu\}@amazon.com} \quad
\texttt{ziyang@cs.jhu.edu} \\[0.6em]
{\small\textsuperscript{*}Equal contribution. \quad
\textsuperscript{\dag}Work conducted during an internship at Amazon. \quad
\textsuperscript{\ddag}Corresponding author.}
}

\begin{document}
\maketitle

\begin{abstract}

Deep research agents are increasingly deployed to retrieve precise information from large corpora, yet multi-hop, long-horizon tasks remain challenging. Synthetic data generation has emerged as a scalable solution, with existing pipelines typically expanding seed tasks to increase their apparent depth and breadth. However, this expansion does not reliably translate into necessary reasoning: newly introduced evidence may be only weakly coupled to the answer, creating a mismatch between intended and empirical difficulty. To address this problem, we introduce GRADS, a graph-guided agentic data synthesis framework with explicit difficulty control. GRADS runs a closed loop of seed selection, evidence expansion, question--answer construction, and verification while maintaining a graph of the evidence and inference dependencies required to derive each answer. At each step, it conditions generation on a relevant subgraph, preserving indispensable inference without accumulating irrelevant context. Experiments across security, software-engineering, and scientific-document domains show that GRADS consistently generates verified tasks with empirically separated difficulty distributions, as measured by independent solvers. Moreover, under matched training budgets, post-training on this data delivers consistent gains at every level: SFT raises the solution-termination rate to over $95\%$ while cutting tool calls and turns by up to 75\%, RL trims tool use by up to a further $19.5\%$, and on our held-out, domain-specific benchmarks, the resulting 4B model matches Sonnet~4 on software engineering and surpasses it on the scientific and security domains.
\end{abstract}

\section{Introduction}
\label{sec:intro}

Specialized knowledge is often distributed across domain-specific record collections that are too large to inspect manually.
Deep search agents have emerged as a promising way to exploit such collections: LLM-based agents that iteratively plan, issue search queries, read retrieved documents, and reason over them to answer complex, multi-hop information needs~\citep{huang2025deep,li2025searcho1,li2025webthinker}.
Post-training, including supervised fine-tuning (SFT) and reinforcement learning (RL), has proven to be effective in teaching models when and how to invoke search tools during reasoning~\citep{jin2025searchr1,song2025r1searcher,zheng2025deepresearcher}.
A rapidly growing body of work demonstrates that synthetic question--answer data is sufficient to train strong deep search agents through SFT and RL~\citep{wu2025webdancer,li2025websailor,gao2025beyond,sun2025simpledeepsearcher,sun2025zerosearch,liu2025webexplorer}.
The quality of such agents is therefore bounded by the quality of the data they are trained on, and in particular by how reliably that data exercises genuine multi-step reasoning.

To make synthetic data sufficiently challenging, most pipelines impose an explicit structure intended to raise the latent inference complexity of each question. 
Representative strategies include depth- and width-based task extension~\citep{shi2026taskcraft}, set-theoretic formalization of
information-seeking tasks~\citep{tao2026webshaper}, graph sampling combined with information obfuscation~\citep{li2025websailor,wei2025browsecomp}, recursive construction of hierarchical research trees~\citep{xia2025open}, and bottom-up composition of single-hop questions~\citep{trivedi2022musique}. 
These methods share a common premise that complexity accumulated during synthesis translates into complexity of the reasoning required at solution time. 
But this premise does not reliably hold. 
A long line of analysis shows that nominally multi-hop questions are frequently answerable through shortcuts or disconnected reasoning that bypasses the intended chain~\citep{min2019compositional,jiang2019avoiding,trivedi2020is}.
When evidence is added to a question without regard to whether it is necessary, it may be only weakly coupled to the answer.
The intended difficulty of the resulting item then diverges from its empirical difficulty, and the difficulty label ceases to be a reliable signal for evaluation or curriculum design. 
This motivates the central questions of this work: can we synthesize data whose difficulty is realized in the solver's reasoning,
not merely claimed? Is such data effective for post-training?

We approach these questions by defining the \emph{evidence graph} of a QA pair, whose nodes are source-grounded claims and whose edges are \emph{necessary} inference steps, so that its depth estimates the reasoning an answer requires. 
Building on this structure, we introduce \textbf{GRADS}, a graph-regulated agentic data synthesis method with controllable difficulty.
GRADS departs from two prevailing strategies. The first piles ever more explicit evidence into a question, extending tasks with additional constraints, entities, or obfuscated clues until they appear harder~\citep{shi2026taskcraft,tao2026webshaper,li2025websailor,liu2025webexplorer}. 
The second builds or samples a knowledge graph to broaden the raw material from which questions are drawn~\citep{chen2025graphgen,lu2025deepdive,xia2025open}. 
Both increase the apparent size of a question, but neither guarantees that the added evidence is required to reach the answer. 
GRADS instead deepens a question within an agentic loop by regulating the structure of its evidence graph. 
It decouples exploration from difficulty control, and is designed to ensure that each question attains its required
reasoning depth.

Applying this idea to real domain knowledge, such as GitHub issue trackers and scientific experiment notes, raises two further challenges. 
First, the underlying corpora are too large to be explored exhaustively, so an agent must locate relevant evidence efficiently.
Second, LLM-synthesized answers are prone to hallucination~\citep{huang2023survey}, and an incorrect answer is more damaging in training data than an easy one. 
To address scale, we clean and cluster the raw records and expose unified local and online search tools to the agents. 
To address correctness, each answer carries a rubric of atomic claims tied to verbatim source spans~\citep{gao2023enabling,min2023factscore}, and verification subagents check every item before acceptance~\citep{dhuliawala2023chain,manakul2023selfcheckgpt,du2023improving}.

We evaluate GRADS on three heterogeneous domains: GitHub issue tracking, experimental particle physics, and security.
For each domain, we construct a representative benchmark and a post-training corpus. 
Compared with three synthesis baselines, GRADS is the only method whose delivered reasoning depth increases monotonically with the requested difficulty, and correspondingly the only one whose
solver scores decrease monotonically from Easy to Hard.
This ordering is preserved across solvers of three capability tiers.
The resulting data is also effective for training: supervised fine-tuning a 4B model on GRADS trajectories yields large gains over the untrained model in every domain and outperforms every baseline at matched sample count and optimizer steps. 
Although SFT accounts for most of the post-training margin, subsequent RL further improves tool use and the model's ability to terminate with an answer, bringing the 4B model to the level of the mid-sized Sonnet~4 solver.

Our contributions are as follows:
(i) We formalize the evidence graph of a QA pair, whose necessary-inference depth serves as both a synthesis-time control and an evaluation-time measure.
(ii) We introduce \textbf{GRADS}, which decouples exploration from difficulty control and pairs every item with a source-grounded rubric and multi-stage verification. 
(iii) Across three real-world domains, GRADS delivers controllable, empirically realized difficulty, and its data yields consistent post-training gains over the base model and competing synthesis methods.

\section{Related Work}
\label{sec:related}

\paragraph{Synthetic Data Generation for Agentic Tasks.}
Synthetic data generation has become the dominant paradigm for training deep research agents.
\citet{shi2026taskcraft} introduce TaskCraft, which expands atomic tasks through depth-based and width-based extensions with incremental validation to produce scalable, multi-tool agentic tasks.
\citet{tao2026webshaper} propose WebShaper, formalizing information-seeking tasks through Knowledge Projections to achieve principled generation with controlled reasoning structures.
\citet{wu2025webdancer} establish an end-to-end pipeline spanning browsing data construction, trajectory sampling, SFT, and RL, scaling both the volume and depth of multi-hop reasoning.
At larger scales, \citet{cen2026webscale} convert pretraining documents into 1.2 million verifiable QA pairs, demonstrating that RL data can reach pretraining-level scale.
\citet{xie2026quest} pair synthetic queries with rubric trees for unified verification across fact-seeking, citation grounding, and report synthesis without human annotation.
\citet{tongyi2025deepresearch} model environments as Prior, Simulated, and Real-world to balance stability, fidelity, and cost.
However, existing pipelines condition generation on isolated seeds without an explicit dependency structure, so expanded evidence may be only weakly coupled to the answer.
GRADS addresses this by maintaining a graph of evidence and inference dependencies throughout synthesis.

\paragraph{Difficulty-Aware Data Selection and Curriculum.}
A complementary line of work recognizes that difficulty is a key axis for data selection and curriculum design.
\citet{hu2026dereason} demonstrate that partitioning STEM reasoning data by reasoning intensity---broad-coverage problems for SFT, difficult problems for RL---outperforms random splitting in decoupled SFT-then-RL training.
\citet{gao2026pcl} further show that prompts of intermediate difficulty (roughly 50\% success rate) are most sample-efficient for RL convergence, while \citet{yang2026select2reason} leverage a difficulty-aware reward model to select only 10\% of long-CoT instruction data with competitive performance.
Yet difficulty is not static:
\citet{zhou2026dare} empirically show that existing estimators become unreliable under policy drift, with estimation error growing as training progresses.
This motivates methods that co-evolve difficulty estimators with training~\citep{zhou2026dare}, adaptively select data clusters via bandit-based self-pacing~\citep{nguyen2026space}, and dynamically re-estimate difficulty within upcoming batches to counteract the Difficulty Shift phenomenon~\citep{wu2025adcl}.
GRADS extends this principle into the synthesis loop itself, conditioning expansion on a dependency subgraph and verifying that each introduced piece of evidence is necessary, thereby producing tasks with empirically separated difficulty distributions.

\paragraph{Datasets and Benchmarks for Deep Research.}
A parallel line of work focuses on the datasets and benchmarks that both evaluate and feed data synthesis for deep research agents.
Expert-written suites such as \citet{wei2025browsecomp} and Humanity's Last Exam~\citep{phan2025hle} push the difficulty ceiling with hard-to-retrieve, short-answer questions, while \citet{chen2026browsecompplus} fix a reproducible corpus to disentangle retrieval from end-to-end agent accuracy.
To reduce reliance on human authoring, \citet{shen2026hopweaver} synthesize authentic bridge and comparison multi-hop questions directly from raw corpora, and \citet{xia2025infoseek} grow Research Trees from large-scale webpages into over 50k hierarchical-constraint tasks with reasoning trajectories.
\citet{chen2026taskevolve} go further by co-evolving queries, DAGs, and rubrics through an Explorer--Formalizer--Challenger loop, yielding verifiable benchmarks whose structure is itself a difficulty signal.
Still, these efforts largely operate over the open web and target generic information-seeking skills; by contrast, GRADS targets specialized, heterogeneous domain corpora (GitHub, physics, security) and treats both the questions and their evidence graphs as artifacts for post-training, not only for evaluation.
\section{Question Expansion with Graph Structure}
\label{sec:graph}

Difficulty control in GRADS rests on an \emph{evidence graph} built online during exploration, over
which a deterministic complexity is computed.
This section defines the graph and its complexity, and
describes the process that grows it; how the agents consume it is deferred to \S\ref{sec:method}.

\subsection{Graph Structure and Definitions}
\label{sec:graph:def}

\begin{definition}[Evidence graph]\label{def:graph}
An evidence graph is $G=(V,E,\tau,\pi,a)$ with $(V,E)$ a DAG:
$V$ is a set of \emph{claims} (atomic facts, one per retrieved record);
$E$ carries \emph{necessary} dependency edges---$(u,v)\in E$ means establishing $v$ requires $u$, so topology proxies reasoning effort, not relatedness;
$\tau:E\to\mathcal{R}=\{\texttt{supports},\texttt{causes},\texttt{refutes},\dots\}$ labels edges; provenance
$\pi:V\to\Sigma$ maps each claim to its source context (with $\Sigma$ the space of source contexts);
and $a\in V$ is a distinguished \emph{anchor}.
$\mathcal{P}(G)$ denotes directed paths ending at $a$, and $|p|$ their edge count.
\end{definition}

\begin{definition}[Node types]\label{def:types}
Each claim is a \emph{core} node (a main conclusion / root cause; may anchor a question), an \emph{inference} node (derived by combining other claims; where multi-hop depth accumulates), or a \emph{detail} node (a supporting fact---key, date, version, status---that decorates a core or inference node but never lies on the reasoning chain).
\end{definition}

\paragraph{Structural functionals.}
Writing $d^{-}_v,d^{+}_v$ for the in-/out-degree of $v$, we compute four quantities in code
(never by an LLM).
\emph{Depth} is the longest necessary chain and \emph{breadth} the number of
independent branches converging on the anchor $a$; \emph{size} is the number of vertices; and
\emph{information flow} sums the squared throughput of intermediate hub nodes:
\begin{equation}\label{eq:functionals}
  d(G)=\!\!\max_{p\in\mathcal{P}(G)}\!\!|p|,\quad
  b(G)=d^{-}_a,\quad
  \mathrm{size}(G)=|V|,\quad
  c_{\mathrm{info}}(G)=\!\!\sum_{v\in V}\!\bigl(d^{-}_v d^{+}_v\bigr)^{2}.
\end{equation}
From these we derive a single \emph{complexity} score, a weighted combination that lets subgraphs be
compared on one scale:
\begin{equation}\label{eq:complexity}
  \mathcal{C}(G)=w_1|V|+w_2\,d(G)+w_3\,b(G) +w_4\,c_{\mathrm{info}}(G),
\end{equation}
with the depth and information-flow weights kept dominant so a deep chain is not mistaken for a
wide-but-shallow one.

\paragraph{Difficulty bands.}
Difficulty is governed by depth.
Given integer thresholds
$\theta_{\mathrm{E}}<\theta_{\mathrm{M}}<\theta_{\mathrm{H}}$,
\begin{equation}\label{eq:bands}
  \mathrm{band}(G)=
  \begin{cases}
    \textsc{Easy}       & d(G)\le\theta_{\mathrm{E}},\\
    \textsc{Medium}     & \theta_{\mathrm{E}}<d(G)\le\theta_{\mathrm{M}},\\
    \textsc{Hard}       & d(G)\ge\theta_{\mathrm{H}},
  \end{cases}
\end{equation}
with depths in $(\theta_{\mathrm{M}},\theta_{\mathrm{H}})$ left as a gap between Medium and Hard, and
breadth held under a per-band cap so the knob moves ``deeper,'' not merely ``wider'' (values in
\S\ref{sec:exp}).

\subsection{Expansion Process}
\label{sec:graph:expand}
\begin{figure}[t]
  \centering
  \includegraphics[width=1.0\textwidth]{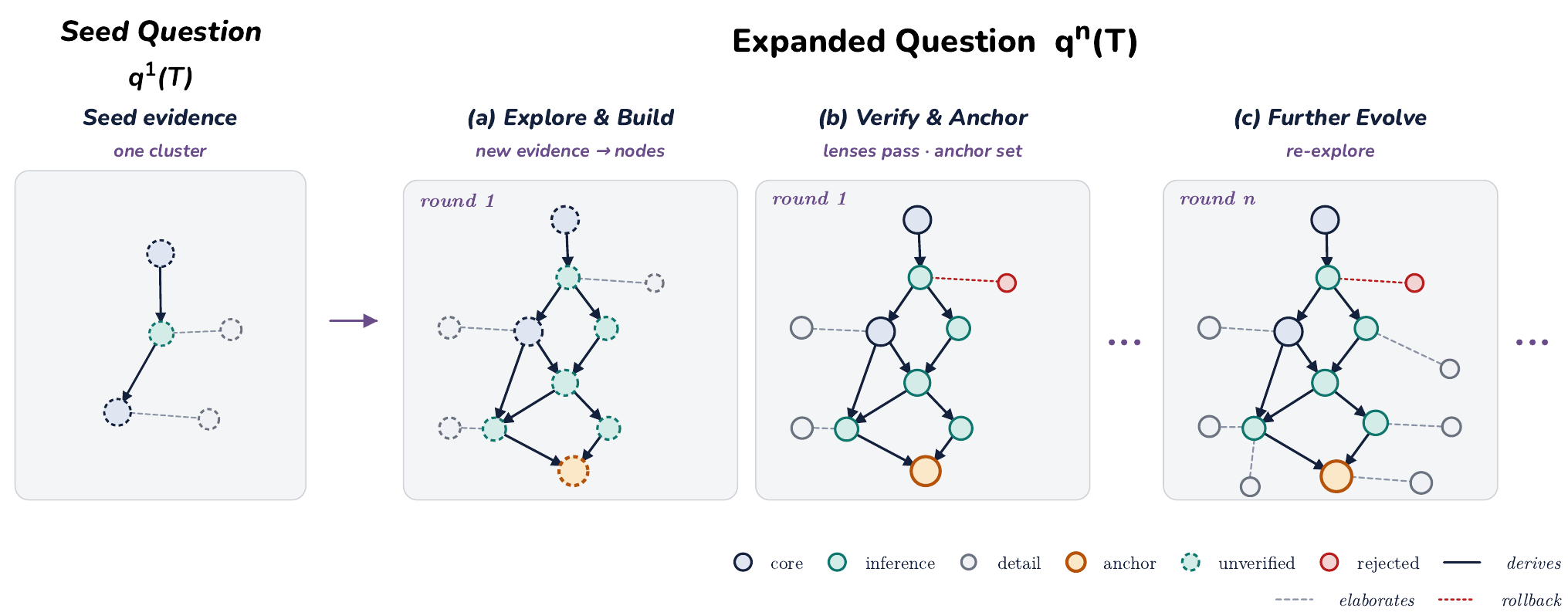}
  \caption{Graph expansion. Exploration adds provenance-linked claim nodes and necessary-dependency
  edges across rounds; reasoning depth accumulates until the requested band is reached, at which
  point a subgraph of the target depth is selected and verbalized into one multi-hop question.}
  \label{fig:expansion}
\end{figure}

In the expansion process, a question is hardened by growing its evidence graph across
rounds (Figure~\ref{fig:expansion}), following a \emph{seed}~$\to$~\emph{explore}~$\to$~\emph{verify}~$\to$~\emph{re-explore} loop.
Let $G_0$ be the seed graph whose claims
already answer a seed question, and let $\theta$ be the depth threshold of the requested band
(Eq.~\ref{eq:bands}).
At round $t$, exploration proposes a set of candidate edits
$\mathcal{K}_t=\{\Delta_t\}$, where each $\Delta_t=(\Delta V^{+},\Delta E^{+};\,\Delta V^{-},\Delta E^{-})$
specifies claims and edges to add and to remove.
A verification operator $\nu$ retains only the necessary part of the additive component, discarding
hops a solver could bypass:
\begin{equation}\label{eq:verify}
  \nu(\Delta_t)=\bigl(\Delta V^{+},\ \{(u,v)\in\Delta E^{+}: v \text{ is underivable without } u\};\ \Delta V^{-},\Delta E^{-}\bigr).
\end{equation}
We then commit the verified edit that most extends the longest necessary chain, so depth grows
rather than breadth:
\begin{equation}\label{eq:expand}
  G_{t+1}=G_t\oplus\nu(\Delta_t^{\star}),\qquad
  \Delta_t^{\star}=\argmax_{\Delta_t\in\mathcal{K}_t}
  \bigl[d\bigl(G_t\oplus\nu(\Delta_t)\bigr)-d(G_t)\bigr],
\end{equation}
where $G\oplus\Delta$ lets one round add, remove, or edit (remove$+$add) claims and edges; ties are broken by
higher $c_{\mathrm{info}}$, then arbitrarily.
The loop re-explores until the first round $T=\min\{t: d(G_t)\ge\theta\}$, at which point $G_T$ is
handed on for selection.
Since a single retrieval rarely yields a deep chain, depth accumulates over
these rounds, and the target $\theta$ the process drives toward and the depth $d(G_T)$ read back off
the final graph are the same quantity---so difficulty is set and measured on one scale.

\section{Method}
\label{sec:method}

\begin{figure}[t]
  \centering
  \includegraphics[width=1.0\textwidth]{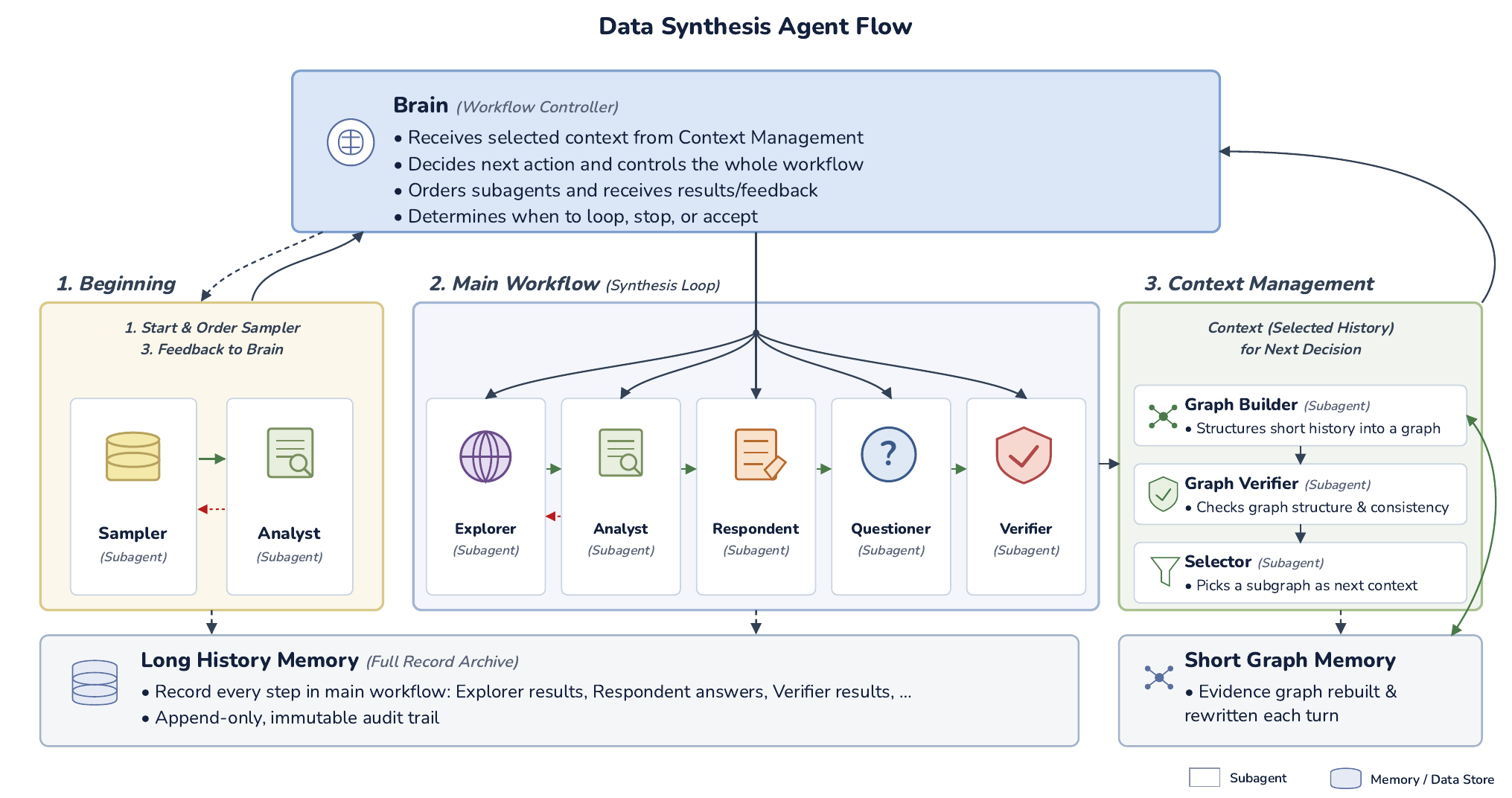}
  \caption{Data synthesis agent flow. A stateless Brain drives
the loop from the selected context. (1) Beginning: the \emph{Sampler}
draws a seed and the \emph{Analyst} sets the generation target.
(2) Main Workflow: each round, \emph{Explorer}, \emph{Analyst},
\emph{Respondent}, \emph{Questioner}, and \emph{Verifier} build and check one
QA item (green $=$ information, red $=$ feedback). (3) Context
Management: the \emph{Graph Builder}, \emph{Graph Verifier}, and \emph{Selector}
maintain the evidence graph and return the target-shaped subgraph as the next
context. An append-only Long History Memory and a rewritten Short
Graph Memory back the loop.}
  \label{fig:main_structure}
\end{figure}

GRADS decouples the exploration process from the difficulty-control process: a main workflow carries
out synthesis, while a context-management stage controls difficulty by tailoring the graph structure
of the context.
The overall process is sketched in Figure~\ref{fig:main_structure}, and we elaborate
each stage in the sections that follow.

\subsection{Data Preprocessing}
\label{sec:method:prep}


We cast each domain's heterogeneous records into a common schema and deduplicate them.
We then partition an induced \emph{relation graph} into clusters, where an edge connects two records that share provenance, such as a common project or artifact.
Seeding each synthesis run from a single cluster bounds its initial exploration to a coherent neighborhood of related evidence and establishes a stable provenance scope.
To support the long inference chains required by multi-hop questions, we provide retrieval tools that iteratively expand the available evidence~\citep{trivedi2023interleaving}.
For each domain, we build two local indices over the clustered corpus: a structured index over record identifiers and metadata for exact lookup, and a BM25 full-text index for ranked retrieval, which remains a strong baseline across heterogeneous domains~\citep{thakur2021beir}.
Document bodies stored outside the metadata, such as attached files, are extracted and cached on first access.
The local indices, online search backend, and auxiliary tools are exposed through a common MCP interface, allowing the same exploration procedure to operate across domains.

\subsection{Agentic Main Loop}
\label{sec:method:loop}

\paragraph{Main workflow.}
A stateless \emph{Brain} orchestrates a role-decomposed team, reading only the context the
context manager selects and choosing each round whether to explore further, generate, or finalize.
A \emph{Sampler} draws the seed once; thereafter an \emph{Explorer} runs the Brain's search order
against the tools and an \emph{Analyst} judges whether the returned evidence advances the target,
while on a generate step a \emph{Respondent} drafts the answer and its rubric and a \emph{Questioner}
turns them into one hard multi-hop question that a \emph{Verifier} then gates.
The Brain interleaves
these explore and generate steps to accumulate reasoning depth, so the subagents pass evidence
forward until the item is verified and on-target; each subagent's precise function and full system
prompt are given in Appendix~\ref{app:prompts}.

\paragraph{Long history memory.}
The Long History is an append-only log with no delete, truncate, or overwrite operation: every seed
paragraph and explored snippet is recorded as one immutable event with a
monotonically increasing id.
Graph nodes never copy this text and store only a bare event id, from
which the original span is recovered on demand by dereferencing.
This keeps the
evidence lossless and verbatim. 

\paragraph{Questions, rubrics, and verification.}
We use ten question types (full taxonomy in Appendix~\ref{app:qtypes}), and for each we build a
rubric of verifiable claims in three layers: \emph{core}, \emph{detail}, and \emph{inference}, as in Definition~\ref{def:types}.
Each claim corresponds to a verifiable source span.
Each item is evaluated in two stages.
A deterministic validator first checks constraints that can be verified programmatically, including
rubric-weight normalization, whether quotations match their cited source spans, and whether references resolve.
Items that pass these checks are then assessed by two adversarially prompted LLM verifiers.
The \emph{provenance verifier}
determines whether each claim is entailed by its cited evidence, while the \emph{reasoning verifier} determines whether each inference step is necessary to derive the answer.
Verification is the crux of the pipeline: since every claim is grounded in a verbatim source span and independently checked, each accepted item carries a per-sample verification record, which provides evidence that the intended reasoning dependencies are present and that its answer is supported by its sources.

\subsection{Context Management}
The short graph memory is the state $G_t=(V_t,E_t,\tau,\pi,a)$ (\S\ref{sec:graph}) at turn $t$,
where difficulty is tied to the complexity of the latent inference graph the answer requires
(verified in \S\ref{sec:exp:synth}).
For every turn, the \emph{Graph Builder}
commits the verified increment of Eq.~\ref{eq:expand}, $G_t=G_{t-1}\oplus\nu(\Delta_{t-1}^{\star})$ with
$\nu$ the necessity filter of Eq.~\ref{eq:verify}, and rewrites the single graph file atomically.
Each node $v$ stores only its provenance pointer $\pi(v)\in\Sigma$, not the span itself, so
$|V_t|$ grows as $O(t)$ while the long history addressed by $\pi$ stays complete and
dereferenceable.
Its topology $\phi(G_t)=\bigl(d(G_t),b(G_t),c_{\mathrm{info}}(G_t)\bigr)$
(Eq.~\ref{eq:functionals}) is a pure function evaluated in code, so the quantity the Brain drives
toward and the quantity read back off $G_t$ are the same $\phi$, not a model's estimate of it.

We use selection to set the difficulty instead of prompting.
Let $a$ be the answer anchor and $\theta$ the depth threshold of the requested band (Eq.~\ref{eq:bands}).
At turn $T$ the \emph{Selector} returns
the subgraph $S^\star$ that becomes the evidence for the next question:
\begin{equation}\label{eq:select}
  S^\star=\argmin_{S\in\mathcal{S}(G_T)}\ \bigl|\,d(S)-\theta\,\bigr|,
  \qquad
  \mathcal{S}(G_T)=\bigl\{\,S\subseteq G_T:\ a\in S,\ \mathrm{Anc}_S(a)\subseteq S,\ b(S)\le\beta\,\bigr\}.
\end{equation}
Each feasible subgraph in $\mathcal{S}(G_T)$ contains the anchor $a$ and all its necessary ancestors $\mathrm{Anc}_S(a)$, and has breadth at most $\beta$.
Among all feasible subgraphs $\mathcal{S}(G_T)$, $S^\star$ is
the one whose reasoning depth $d(S)$ is closest to the target $\theta$.
Only the dereferenced context $\pi(S^\star)$ is passed back to the \emph{Brain}, and the
next question is composed against exactly this context, so its realized structure obeys
$\mathrm{band}(S^\star)$: selection turns the scalar target $\theta$ into a hard constraint on
generation.
Since one retrieval rarely reaches $\theta$, depth accumulates across rounds by
interleaving exploration and generation.

An item is \emph{delivered} only when it passes verification and its selected subgraph realizes exactly the requested band, with the \emph{Graph Verifier} ensuring every node and edge is faithful to
its source. 
Necessity of the chain is enforced by a deletion test: an edge $(u,v)$ counts toward depth only if $v$ cannot be derived without $u$, and otherwise it is kept as a non-necessary link. 
The \emph{Graph Builder} applies this test when adding edges and never lets a detail node carry the chain; the Graph Verifier re-applies it and merges near-duplicate nodes that would lengthen a chain without adding reasoning. 
Since $d(G)$ is computed only over necessary edges and the \emph{Selector} keeps subgraphs closed under necessary ancestors, the delivered depth is designed to count only hops a solver cannot bypass, which the \emph{Verifier}'s reasoning lens checks once more on the finished item.

\section{Experiments}
\label{sec:exp}


\subsection{Domains and Setup}
\label{sec:exp:domains}
We apply GRADS to three heterogeneous corpora of real-world records spanning distinct domains.
\textbf{GitHub} contains issues, comments, and linked JIRA tickets from Apache \texttt{airflow}, \texttt{beam}, and so on (raw size: $7.6$\,GB across ${\sim}1.07$M records).
\textbf{Physics} contains internal analysis notes and public physics analysis summaries from the Compact Muon Solenoid (CMS), a general-purpose particle detector at CERN's Large Hadron Collider (raw size: $4.9$\,GB across ${\sim}5.9$k files).
\textbf{Security} contains red-teaming agent traces from cybersecurity web-application attack and injection challenges (raw size: $1.3$\,GB across ${\sim}18.8$k files).
Synthesis and judging use the Opus~4.8 API, and empirical difficulty is measured by a \emph{solver}---an unmodified agent with the domain tools, a $25$-turn cap, and temperature $0.7$---whose answers are scored against each item's rubric.
For post-training, we fine-tune \textbf{Qwen3.5-4B}~\citep{qwen3} and then run GRPO~\citep{grpo} from the SFT checkpoint, with the full training configuration in Appendix~\ref{app:posttrain}.

\subsection{Data-Synthesis Main Results}
\label{sec:exp:synth}
\begin{figure}[t]
  \centering
  \includegraphics[width=\textwidth]{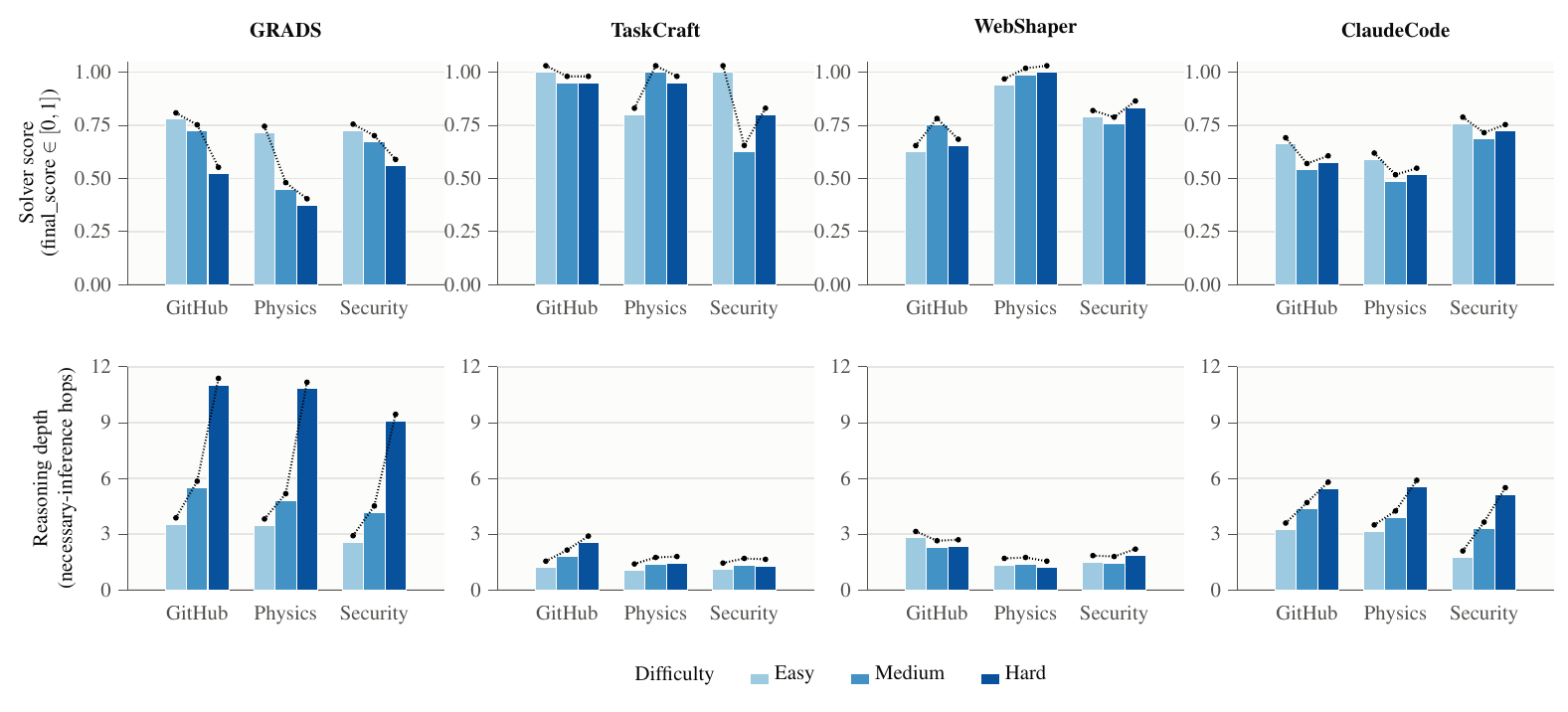}
  \caption{Difficulty is controllable and empirical only for GRADS: delivered depth (bottom row) and
  solver score (top row) by requested band. For more details, check Appendix Table~\ref{tab:app-comparison}.}
  \label{fig:controllability}
\end{figure}

We compare GRADS against three synthesis methods run through an identical protocol: \textbf{Claude-Code}, natural-language prompting for a target difficulty (a prompt-only ablation of GRADS with the graph and Selector removed); 
\textbf{TaskCraft}~\citep{shi2026taskcraft}, which scales task complexity through straightforward extension in depth and width; 
and \textbf{WebShaper}~\citep{tao2026webshaper}, which expands set-theoretic task formalizations. 
Figure~\ref{fig:controllability} reports delivered reasoning depth (bottom row) and solver score (top row) by requested band. Reported depth is measured on each item's full evidence graph, whereas GRADS controls only the selected subgraph, so average depths can exceed the band thresholds.

\paragraph{Depth as the difficulty axis.}
The complexity score $\mathcal{C}(G)$ (Eq.~\ref{eq:complexity}) combines several structural axes, but they are not equally informative about how hard an item is for a solver. 
Measured against rubric-graded solver score, depth $d(G)$ shows the strongest negative correlation (pooled Spearman $r_s=-0.45$), ahead of information flow $c_{\mathrm{info}}(G)$ ($-0.39$) and graph size $|V|$ ($-0.38$), while breadth $b(G)$ is only weakly related ($-0.17$). 
Depth also tracks the other axes closely, so a single depth value captures most of what the full topology encodes.
We therefore band difficulty by $d(G)$ alone; per-domain correlations are in Appendix~\ref{app:depth-complexity}.

\paragraph{Controllable depth predicts empirical difficulty.}
The Easy/Medium/Hard division is applied identically to every method, each under its own reasoning-depth measure: a question of depth no more than $3$ is Easy, depth $4$--$5$ is Medium, and depth $\ge 8$ is Hard.
The bottom row of Figure~\ref{fig:controllability} shows that straightforward accumulation of evidence often fails to add the necessary inference a question requires, so a method can retrieve more yet not grow deeper.
GRADS is the only method whose delivered depth rises monotonically, and correspondingly the only one whose solver scores fall monotonically (top row).
TaskCraft~\citep{shi2026taskcraft} and WebShaper~\citep{tao2026webshaper} instead stay flat under our measure, so their solver scores also stay flat across bands.
Claude-Code deepens its questions steadily with the language prompt but never reaches the required depth, and its Medium and Hard scores do not separate.

\subsection{Post-Training: SFT Main Results}
\label{sec:exp:sft}

\begin{figure}[t]
  \centering
  \includegraphics[width=\textwidth]{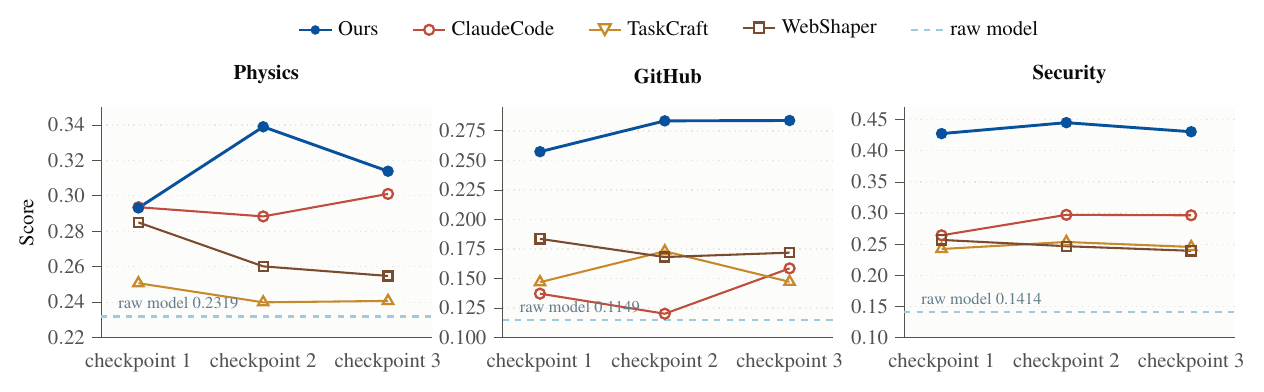}
  \caption{SFT on GRADS data (Ours) dominates every synthesis baseline in every domain; dashed line is the raw model. 
  Check more details in Appendix Table~\ref{tab:app-sftckpt}.}
  \label{fig:sft}
\end{figure}

From each delivered QA item, we reconstruct agentic trajectories that replay the retrieval-and-reasoning path grounding the answer (construction details in Appendix~\ref{app:posttrain}), and fine-tune Qwen3.5-4B on these trajectories. 
Each baseline's QA is consumed by the same builder, configuration, and benchmark, so only the data differs. 
All arms are evaluated on the same frozen benchmark (Figure~\ref{fig:sft}). 
On every domain, SFT on GRADS trajectories beats the raw model by a significant margin and beats every baseline arm, supporting the utility of the synthesized data.

\subsection{Post-Training: RL Main Results}
\label{sec:exp:rl}
\begin{figure}[t]
  \centering
  \begin{minipage}{0.40\textwidth}
    \centering
    \includegraphics[width=\textwidth]{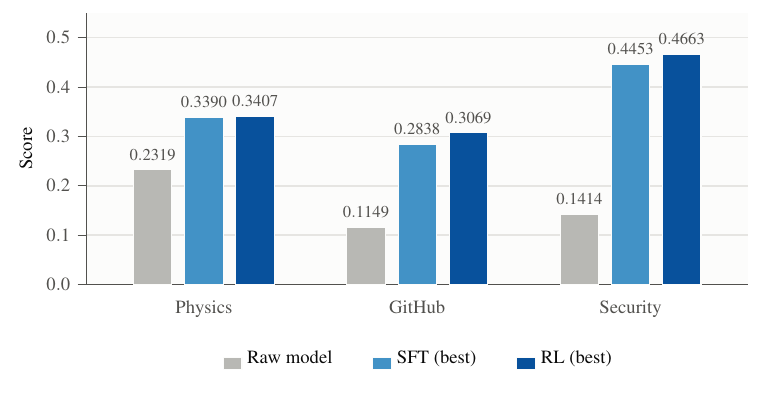}
    \caption{Post-training performance. SFT accounts for most of the score improvement, while RL provides modest additional gains. Check more details in Appendix Table~\ref{tab:app-rl}.}
    \label{fig:utility}
  \end{minipage}\hfill
  \begin{minipage}{0.59\textwidth}
    \centering
    \includegraphics[width=\textwidth]{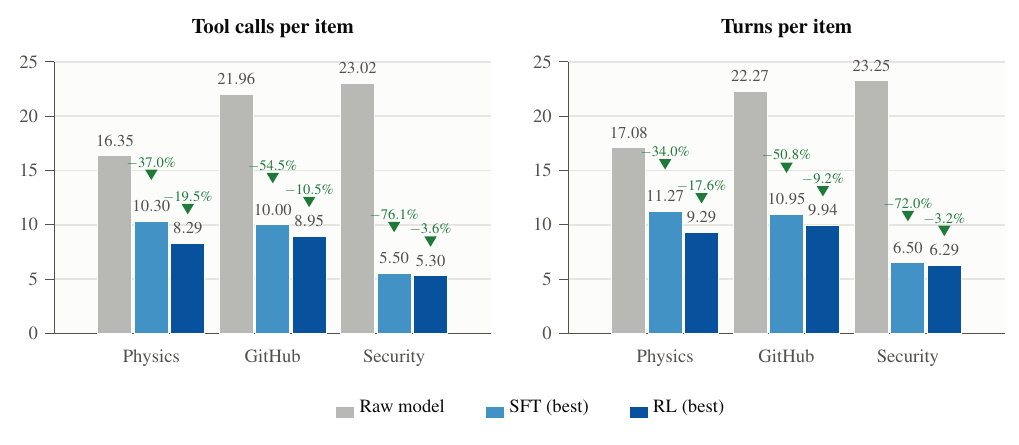}
    \caption{Tool calls and turns per item: SFT cuts both sharply, RL trims a little more. Check more details in Appendix Table~\ref{tab:app-behaviour}.}
    \label{fig:tool}
  \end{minipage}
\end{figure}

We further apply standard GRPO~\citep{grpo} to each domain's SFT checkpoint, training on $240$ synthesized pairs disjoint from the benchmark. 
Each rollout is rewarded by the same claim-level judge that scores the benchmark, but applied to the weighted reference claims of its own training pair, plus a bonus for the expected answer format.
We train for $30$ steps (Appendix~\ref{app:posttrain}).
GRPO produces small positive point estimates in all three domains(Figure~\ref{fig:utility}), but none is statistically separable from the SFT policy under our 30-step budget. 
Its clearer effect is improved tool efficiency, reducing tool calls by up to $19.5\%$.
After post-training, the 4B model approaches Sonnet~4 on the GitHub benchmark and surpasses it on the Physics and Security benchmarks (Table~\ref{tab:app-sonnet}).

\subsection{Discussions}
\label{sec:exp:disc}

\begin{figure}[t]
  \centering
  \includegraphics[width=0.95\textwidth]{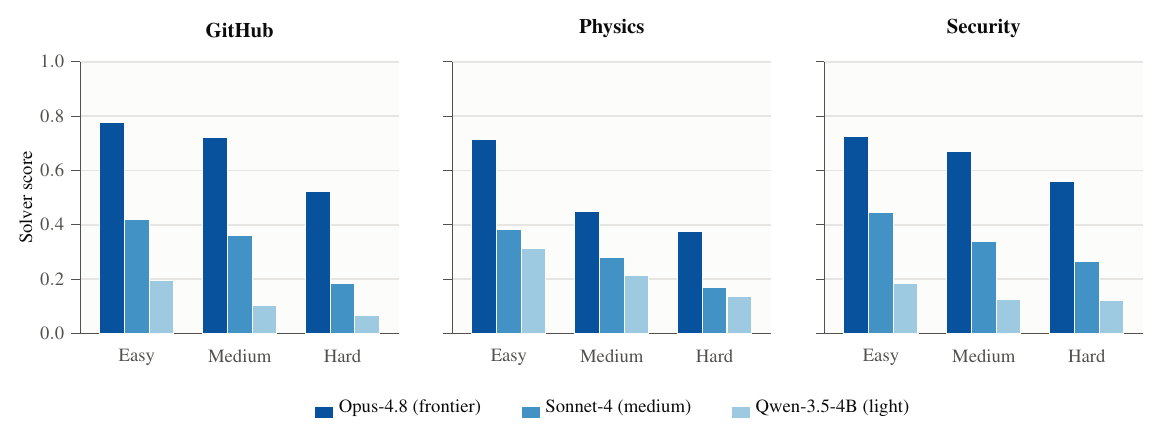}
  \caption{
    Every capability tier degrades from Easy$\to$Hard; capability is a level shift.
    For more details, check Appendix Table~\ref{tab:app-ladder}.
  }
  \label{fig:ladder}
\end{figure}

\paragraph{The difficulty axis is model-agnostic.}
To eliminate the influence of any single model, we evaluate three capability tiers: Opus~4.8 (frontier), Sonnet~4 (medium), and Qwen3.5-4B (light). 
Across all three domains, the solver score decreases monotonically as the model gets weaker, and for each model it decreases as the problems get harder (Figure~\ref{fig:ladder}).
These results suggest that difficulty ordering is a property of the data, not the solver.

\paragraph{Tool analysis.}
SFT provides the largest improvement in tool-use and termination behavior (Figure~\ref{fig:tool}): across all three domains the automatic-termination rate, measured by the presence of a boxed final answer, increases from $6\%$ to above $95\%$. 
Tool calls and turns per item are roughly halved to quartered (Figure~\ref{fig:tool}), and the rate of items hitting the $25$-turn cap falls from $28$--$77\%$ to $0$--$5\%$. 
Although the termination rate remains nearly unchanged after RL, GRPO consistently reduces tool calls and turns further, by up to $19.5\%$ relative to the SFT results.

\paragraph{Cost analysis.}
Synthesis cost, measured in API calls and tokens per delivered item, scales with the requested difficulty because a deeper chain needs more explore--generate rounds.
A Hard item takes ${\approx}2.3\times$ the calls and ${\approx}2.4\times$ the tokens of an Easy one (Appendix~\ref{app:cost}, Table~\ref{tab:app-cost}).
Compared with baselines (Table~\ref{tab:app-costcmp}), GRADS strikes the best balance of cost and results.

\subsection{Limitations}
First, our evidence that the synthesized data is \emph{useful} for post-training is confined to supervised fine-tuning: under a $30$-step budget, GRPO did not yield a separable improvement over the SFT policy (\S\ref{sec:exp:rl}), so whether reinforcement learning adds value under a larger budget remains untested.
Second, because the Selector targets the requested depth by assembling a necessary-ancestor subgraph, in some domains it admits additional context that pushes an item's realized depth beyond the band threshold (Figure~\ref{fig:controllability}, Table~\ref{tab:app-comparison}); this overshoot is benign for difficulty but may introduce a mild bias in the delivered depth distribution.
Third, difficulty control is reliable at the level of the three bands, but the exact reasoning depth of any single item is noisier, so per-item difficulty should be treated as approximate rather than exact, and this residual variation differs across solvers.
Finally, the same evidence graph both defines an item's difficulty and supplies the subgraph the question is built from, so difficulty is in part validated by the construction that produced it. Moreover, whether an edge is necessary is judged by LLM agents (deletion tests by the Graph Builder and Graph Verifier, and the reasoning lens) rather than by human annotators, so these checks provide evidence, not proof, that a solver cannot bypass a hop; a human audit of edge necessity is left to future work.

\section{Conclusion}
\label{sec:conclusion}
We presented GRADS, an agent that synthesizes controllable and verifiable QA data from
heterogeneous real-record corpora. Its central idea is to decouple open-ended exploration from
deterministic difficulty control. Exploration grows an evidence graph, and difficulty is set by a
graph-complexity level computed over that graph: reasoning depth is the governing axis, together
with breadth, size, and information flow through hub nodes. Complexity is therefore a control
knob at synthesis time and a measured quantity at evaluation time, and every answer claim stays
traceable to a source span.
This design yields three results. First, the requested and the achieved complexity levels align
closely. Second, the induced difficulty is empirical: solver accuracy degrades monotonically with
the requested level, and this trend holds under an independent judge and an independent graph
builder. Third, the data trains, since SFT on GRADS trajectories improves a 4B model substantially,
by up to roughly a threefold gain on the strongest domain, and beats three synthesis baselines at
matched sample count. The same controlled, verified corpus thus serves both evaluation and post-training.


\section{Acknowledgements}
We thank Amazon for their generous support and resources, and the Amazon Nova AI Challenge for providing the dataset used in this work. We are also grateful to our advisors and colleagues for their guidance, feedback on earlier drafts, and insightful discussions throughout this project.

\newpage

\section*{Reproducibility Statement}

We release the agent source code, together with the agent prompts and evaluation rubrics that define the pipeline. 
All comparisons use a fixed benchmark/solver/judge with paired bootstrap and permutation tests at fixed seeds.

\section*{Ethics Statement}

All source records are used under their original licenses, and any sensitive content is handled in line with responsible-disclosure norms. 
Because source documents may appear in model pretraining, we report benchmark scores as relative comparisons on a fixed harness rather than contamination-free absolutes.

\section*{AI Use Statement}


Generative AI is integral to our method: GRADS uses large language models as the agents that explore corpora, synthesize question--answer pairs, and verify them (Section~\ref{sec:method}). Otherwise, we used generative AI only to assist with editing for readability and with auxiliary code such as plotting and data-processing scripts. All AI-assisted output was reviewed by the authors, who take full responsibility for the final content.




\bibliographystyle{iclr2027_conference}
\bibliography{references}

\appendix
\newpage
\section{Numerical data for figures}
\label{app:data}

This appendix reports the exact values behind every figure in Section~\ref{sec:exp}. The
\emph{Physics} domain is the CMS experimental-physics notes (on-disk name \texttt{cms}).

\begin{table}[ht]
  \centering
  \caption{Data behind Figure~\ref{fig:controllability}. Solver score (rubric-weighted,
  $\in[0,1]$; $n=20$ per cell) and delivered reasoning depth (longest necessary-inference chain) by
  method $\times$ domain $\times$ requested band. Only GRADS shows a monotone Easy$>$Medium$>$Hard
  solver-score gradient together with depth that grows Easy$\to$Hard.}
  \label{tab:app-comparison}
  \begin{tabular}{llccc@{\hskip 2em}ccc}
    \toprule
    & & \multicolumn{3}{c}{Solver score} & \multicolumn{3}{c}{Reasoning depth} \\
    \cmidrule(lr){3-5}\cmidrule(lr){6-8}
    Method & Domain & Easy & Medium & Hard & Easy & Medium & Hard \\
    \midrule
    \multirow{3}{*}{\textbf{GRADS}}
      & GitHub   & 0.778 & 0.722 & 0.522 & 3.53 & 5.50 & 11.03 \\
      & Physics      & 0.715 & 0.450 & 0.374 & 3.47 & 4.83 & 10.82 \\
      & Security & 0.725 & 0.671 & 0.559 & 2.57 & 4.17 & 9.10 \\
    \midrule
    \multirow{3}{*}{TaskCraft}
      & GitHub   & 1.000 & 0.950 & 0.950 & 1.20 & 1.80 & 2.55 \\
      & Physics      & 0.800 & 1.000 & 0.950 & 1.05 & 1.40 & 1.45 \\
      & Security & 1.000 & 0.625 & 0.800 & 1.10 & 1.35 & 1.30 \\
    \midrule
    \multirow{3}{*}{WebShaper}
      & GitHub   & 0.624 & 0.752 & 0.654 & 2.80 & 2.30 & 2.35 \\
      & Physics      & 0.938 & 0.988 & 1.000 & 1.35 & 1.40 & 1.20 \\
      & Security & 0.789 & 0.758 & 0.834 & 1.50 & 1.45 & 1.85 \\
    \midrule
    \multirow{3}{*}{ClaudeCode}
      & GitHub   & 0.661 & 0.540 & 0.576 & 3.25 & 4.35 & 5.45 \\
      & Physics      & 0.589 & 0.487 & 0.518 & 3.15 & 3.90 & 5.55 \\
      & Security & 0.758 & 0.685 & 0.723 & 1.75 & 3.30 & 5.15 \\
    \bottomrule
  \end{tabular}
\end{table}

\subsection{Reasoning depth as the dominant complexity axis}
\label{app:depth-complexity}

We report the per-item correlation between rubric-graded solver score and each code-measured
topology axis, on the delivered benchmarks (GitHub $n=100$, Physics $n=100$, Security $n=135$; pooled
$n=335$; solver $=$ DeepAgent on Opus~4.8, single sample, score $\in[0,1]$). Depth is the axis most strongly (negatively) correlated with solver score on GitHub, Physics, and
pooled; on Security, size and dispersion are marginally stronger. Depth also tracks the other axes
closely (depth--$c_{\mathrm{info}}$ and depth--size Spearman $r_s\approx0.6$--$0.85$),
so a single depth scalar captures most of what the full topology encodes. Pearson and Spearman
coefficients agree, indicating a monotone, approximately linear negative trend; the linear fit is
moderate ($R^2\approx0.2$--$0.3$ for depth), as expected since structural complexity is a controlled
difficulty variable, not a complete determinant of success. This justifies governing the difficulty
bands by depth.

\begin{table}[ht]
  \centering
  \caption{Spearman $r_s$ between rubric-graded solver score and each term of $\mathcal{C}(G)$, per
  domain and pooled. Depth is the strongest predictor on GitHub, Physics, and pooled; dispersion is
  the number of distinct source records.}
  \label{tab:app-depthcomplexity}
  \begin{tabular}{llcccc}
    \toprule
    Axis & Symbol & GitHub & Physics & Security & Pooled \\
    \midrule
    Depth            & $d(G)$              & $-0.50$ & $-0.51$ & $-0.49$ & $-0.45$ \\
    Information flow & $c_{\mathrm{info}}(G)$ & $-0.38$ & $-0.36$ & $-0.39$ & $-0.39$ \\
    Size             & $|V|$               & $-0.26$ & $-0.26$ & $-0.35$ & $-0.38$ \\
    Breadth          & $b(G)$              & $-0.12$ & $-0.07$ & $-0.10$ & $-0.17$ \\
    Dispersion       & --                  & $-0.44$ & $-0.34$ & $-0.44$ & $-0.38$ \\
    \bottomrule
  \end{tabular}
\end{table}
\begin{table}[ht]
  \centering
  \caption{SFT benchmark score at three checkpoints per arm (step in parentheses;
  best in bold), with the best score and its change versus the raw model and versus ours. Ours is the
  best arm in every domain.}
  \label{tab:app-sft}
  \small
  \setlength{\tabcolsep}{4pt}
  \begin{tabular}{llccccccc}
    \toprule
    Domain & Arm & Ckpt 1 & Ckpt 2 & Ckpt 3 & Best & $\Delta$ raw & $\Delta$ ours \\
    \midrule
    \multirow{5}{*}{Physics}
      & raw Qwen3.5-4B & \multicolumn{3}{c}{--} & 0.2319 & -- & -- \\
      & \textbf{Ours} & 0.2932\,{\scriptsize(10)} & \textbf{0.3390}\,{\scriptsize(20)} & 0.3139\,{\scriptsize(24)} & \textbf{0.3390} & $+0.1071$ & -- \\
      & Claude-Code & 0.2936\,{\scriptsize(12)} & 0.2884\,{\scriptsize(20)} & \textbf{0.3011}\,{\scriptsize(24)} & 0.3011 & $+0.0692$ & $-0.0379$ \\
      & TaskCraft & \textbf{0.2507}\,{\scriptsize(12)} & 0.2399\,{\scriptsize(20)} & 0.2407\,{\scriptsize(24)} & 0.2507 & $+0.0188$ & $-0.0883$ \\
      & WebShaper & \textbf{0.2850}\,{\scriptsize(12)} & 0.2602\,{\scriptsize(16)} & 0.2548\,{\scriptsize(20)} & 0.2850 & $+0.0531$ & $-0.0540$ \\
    \midrule
    \multirow{5}{*}{GitHub}
      & raw Qwen3.5-4B & \multicolumn{3}{c}{--} & 0.1149 & -- & -- \\
      & \textbf{Ours} & 0.2574\,{\scriptsize(8)} & 0.2835\,{\scriptsize(16)} & \textbf{0.2838}\,{\scriptsize(26)} & \textbf{0.2838} & $+0.1689$ & -- \\
      & Claude-Code & 0.1372\,{\scriptsize(12)} & 0.1202\,{\scriptsize(20)} & \textbf{0.1586}\,{\scriptsize(24)} & 0.1586 & $+0.0437$ & $-0.1252$ \\
      & TaskCraft & 0.1469\,{\scriptsize(16)} & \textbf{0.1731}\,{\scriptsize(20)} & 0.1472\,{\scriptsize(24)} & 0.1731 & $+0.0582$ & $-0.1107$ \\
      & WebShaper & \textbf{0.1836}\,{\scriptsize(16)} & 0.1682\,{\scriptsize(20)} & 0.1719\,{\scriptsize(24)} & 0.1836 & $+0.0687$ & $-0.1002$ \\
    \midrule
    \multirow{5}{*}{Security}
      & raw Qwen3.5-4B & \multicolumn{3}{c}{--} & 0.1414 & -- & -- \\
      & \textbf{Ours} & 0.4277\,{\scriptsize(16)} & \textbf{0.4453}\,{\scriptsize(20)} & 0.4307\,{\scriptsize(24)} & \textbf{0.4453} & $+0.3039$ & -- \\
      & Claude-Code & 0.2643\,{\scriptsize(12)} & \textbf{0.2972}\,{\scriptsize(20)} & 0.2965\,{\scriptsize(24)} & 0.2972 & $+0.1558$ & $-0.1481$ \\
      & TaskCraft & 0.2423\,{\scriptsize(12)} & \textbf{0.2536}\,{\scriptsize(20)} & 0.2456\,{\scriptsize(24)} & 0.2536 & $+0.1122$ & $-0.1917$ \\
      & WebShaper & \textbf{0.2569}\,{\scriptsize(8)} & 0.2468\,{\scriptsize(12)} & 0.2394\,{\scriptsize(16)} & 0.2569 & $+0.1155$ & $-0.1884$ \\
    \bottomrule
  \end{tabular}
\end{table}

\begin{table}[ht]
  \centering
  \caption{Data behind Figure~\ref{fig:ladder}. Mean solver score ($\in[0,1]$, claim-weighted judge
  held constant at Opus~4.8) by model $\times$ domain $\times$ band, for a frontier (Opus~4.8),
  medium (Sonnet~4), and light (Qwen3.5-4B) solver. Every model degrades monotonically
  Easy$>$Medium$>$Hard in every domain.}
  \label{tab:app-ladder}
  \begin{tabular}{llccc}
    \toprule
    Model & Domain & Easy & Medium & Hard \\
    \midrule
    \multirow{3}{*}{Opus-4.8 (frontier)}
      & GitHub   & 0.778 & 0.722 & 0.522 \\
      & Physics  & 0.715 & 0.450 & 0.374 \\
      & Security & 0.725 & 0.671 & 0.559 \\
    \midrule
    \multirow{3}{*}{Sonnet-4 (medium)}
      & GitHub   & 0.421 & 0.362 & 0.183 \\
      & Physics  & 0.381 & 0.279 & 0.170 \\
      & Security & 0.446 & 0.340 & 0.266 \\
    \midrule
    \multirow{3}{*}{Qwen3.5-4B (light)}
      & GitHub   & 0.194 & 0.102 & 0.065 \\
      & Physics  & 0.313 & 0.214 & 0.138 \\
      & Security & 0.185 & 0.126 & 0.120 \\
    \bottomrule
  \end{tabular}
\end{table}

\begin{table}[ht]
  \centering
  \caption{Data behind Figure~\ref{fig:sft}. Benchmark score at three checkpoints per arm and
  domain (dash where an arm has fewer saved points). Raw Qwen3.5-4B: Physics $0.2319$, GitHub $0.1149$,
  Security $0.1414$.}
  \label{tab:app-sftckpt}
  \begin{tabular}{llccc}
    \toprule
    Domain & Arm & Ckpt 1 & Ckpt 2 & Ckpt 3 \\
    \midrule
    \multirow{4}{*}{Physics}
      & \textbf{Ours}  & 0.2932 & 0.3390 & 0.3139 \\
      & ClaudeCode     & 0.2936 & 0.2884 & 0.3011 \\
      & TaskCraft      & 0.2507 & 0.2399 & 0.2407 \\
      & WebShaper      & 0.2850 & 0.2602 & 0.2548 \\
    \midrule
    \multirow{4}{*}{GitHub}
      & \textbf{Ours}  & 0.2574 & 0.2835 & 0.2838 \\
      & ClaudeCode     & 0.1372 & 0.1202 & 0.1586 \\
      & TaskCraft      & 0.1469 & 0.1731 & 0.1472 \\
      & WebShaper      & 0.1836 & 0.1682 & 0.1719 \\
    \midrule
    \multirow{4}{*}{Security}
      & \textbf{Ours}  & 0.4277 & 0.4453 & 0.4307 \\
      & ClaudeCode     & 0.2643 & 0.2972 & 0.2965 \\
      & TaskCraft      & 0.2423 & 0.2536 & 0.2456 \\
      & WebShaper      & 0.2569 & 0.2468 & 0.2394 \\
    \bottomrule
  \end{tabular}
\end{table}

\begin{table}[ht]
  \centering
  \caption{GRPO benchmark score at four checkpoints (steps $9/18/24/30$), the SFT starting policy,
  and the raw model, behind Figure~\ref{fig:utility}. Best-RL vs.\ best-SFT deltas all fall inside
  the $\pm0.043$ noise floor.}
  \label{tab:app-rl}
  \begin{tabular}{lcccccc}
    \toprule
    Domain & step 9 & step 18 & step 24 & step 30 & SFT start & raw \\
    \midrule
    Physics      & 0.3130 & 0.3226 & 0.3185 & 0.3407 & 0.3390 & 0.2319 \\
    GitHub   & 0.2540 & 0.2732 & 0.2855 & 0.3069 & 0.2835 & 0.1149 \\
    Security & 0.4119 & 0.4488 & 0.4663 & 0.4543 & 0.4277 & 0.1414 \\
    \bottomrule
  \end{tabular}
\end{table}

\begin{table}[ht]
  \centering
  \caption{Post-training closes the capability gap to a mid-sized solver. Benchmark mean score
  ($30$/$30$/$40$-weighted) of the Sonnet-4 solver (from Table~\ref{tab:app-ladder}) versus the
  post-trained Qwen3.5-4B (best RL checkpoint, Table~\ref{tab:app-rl}). After post-training the 4B
  matches Sonnet-4 on GitHub and surpasses it on Physics and Security.}
  \label{tab:app-sonnet}
  \begin{tabular}{lccc}
    \toprule
    Model & GitHub & Physics & Security \\
    \midrule
    Sonnet-4 (mid solver)        & 0.308 & 0.266 & 0.342 \\
    \textbf{Post-trained 4B (best RL)} & \textbf{0.307} & \textbf{0.341} & \textbf{0.466} \\
    \midrule
    $\Delta$ (4B $-$ Sonnet-4)   & $-0.001$ & $+0.075$ & $+0.124$ \\
    \bottomrule
  \end{tabular}
\end{table}

\begin{table}[ht]
  \centering
  \caption{Data behind Figure~\ref{fig:tool} and the tool analysis. Per-item means for the
  raw model, best SFT, and best RL checkpoint; \texttt{\textbackslash boxed\{\}} count and
  items-at-25-turn-cap are counts out of 100. SFT's invariant effect is the jump in boxed rate and
  the drop in tool calls/turns; answer length moves in domain-specific directions.}
  \label{tab:app-behaviour}
  \begin{tabular}{llccccc}
    \toprule
    Domain & Model & Tool calls & Turns & \texttt{boxed} & Ans.\ length & Cap-hits \\
    \midrule
    \multirow{3}{*}{Physics}
      & raw & 16.35 & 17.08 & 6   & 6662 & 28 \\
      & SFT & 10.30 & 11.27 & 97  & 5528 & 4  \\
      & RL  & 8.29  & 9.29  & 100 & 6656 & -- \\
    \midrule
    \multirow{3}{*}{GitHub}
      & raw & 21.96 & 22.27 & 5   & 3172 & 69 \\
      & SFT & 10.00 & 10.95 & 95  & 5776 & 5  \\
      & RL  & 8.95  & 9.94  & 99  & 5465 & -- \\
    \midrule
    \multirow{3}{*}{Security}
      & raw & 23.02 & 23.25 & 5   & 3061 & 77 \\
      & SFT & 5.50  & 6.50  & 99  & 4282 & 0  \\
      & RL  & 5.30  & 6.29  & 98  & 4308 & -- \\
    \bottomrule
  \end{tabular}
\end{table}

\section{Synthesis cost}
\label{app:cost}

Table~\ref{tab:app-cost} reports GRADS synthesis cost per delivered item by requested difficulty:
both API calls and tokens rise monotonically with depth, since a deeper chain requires more
explore--generate rounds. Table~\ref{tab:app-costcmp} compares GRADS to the baselines at the
method level. Read against delivered depth, GRADS spends a moderate budget that the baselines either undercut
only by staying shallow (WebShaper) or exceed without getting deeper (TaskCraft); measurement
quality differs by method, as noted in the table caption.

\begin{table}[ht]
  \centering
  \caption{GRADS synthesis cost per delivered item, by requested difficulty (Opus~4.8). Both API
  calls and tokens rise monotonically with depth.}
  \label{tab:app-cost}
  \begin{tabular}{lcccc}
    \toprule
    Band & API calls & Input tok & Output tok & Total tok \\
    \midrule
    Easy   & $17.5$ & $149$\,k & $31$\,k & $180$\,k \\
    Medium & $30.2$ & $271$\,k & $56$\,k & $327$\,k \\
    Hard   & $40.3$ & $358$\,k & $76$\,k & $434$\,k \\
    \bottomrule
  \end{tabular}
\end{table}

\begin{table}[ht]
  \centering
  \caption{Synthesis cost per item, GRADS versus baselines. Cost is only meaningful alongside
  delivered depth (Table~\ref{tab:app-comparison}): WebShaper is cheapest because it barely explores
  (and stays shallow), TaskCraft is by far the most expensive yet still shallow, and Claude-Code
  spends about as much as GRADS but never reaches the target depth. Measurement notes: GRADS
  (Opus~4.8) and Claude-Code are exact; TaskCraft calls are exact but its tokens are a \emph{floor}
  (metered on ReAct agent-steps only, ${\sim}3.5$k of ${\sim}25.6$k calls, so true tokens are
  higher); WebShaper calls are exact but its tokens are estimated.}
  \label{tab:app-costcmp}
  \begin{tabular}{lccc}
    \toprule
    Method & API calls / item & Tokens / item \\
    \midrule
    \textbf{GRADS}  & \textbf{28} & \textbf{$302$\,k} \\
    Claude-Code     & $33.7$      & $236$\,k \\
    TaskCraft       & ${\sim}278$ & ${>}0.84$\,M (floor) \\
    WebShaper       & ${\sim}2.9$ & ${\sim}22$\,k (est.) \\
    \bottomrule
  \end{tabular}
\end{table}

\newpage
\section{Agent prompts}
\label{app:prompts}

GRADS is a role-decomposed agent team; each role is invoked with a fixed system prompt and a JSON
output schema enforced by a forced tool call, so no role returns free-form prose that must be
parsed. We reproduce the verbatim system prompts below (as used in \texttt{roles.py}). A design
invariant across all of them is that \emph{no prompt asks a model to compute a sum, count claims,
measure depth, or check a substring}: every such quantity is computed in code (\S\ref{sec:method}),
which is what keeps difficulty a measured rather than a self-reported quantity. The
\texttt{GROUNDING} block below is appended to the Analyst (instruction) and Respondent prompts.

\lstset{style=prompt}

\paragraph{Shared grounding block.} Appended to the Analyst (instruction) and Respondent prompts; it
forbids leaking generation scaffolding or container ids and requires every statement to be grounded
only in the provided evidence.
\begin{lstlisting}
SOURCE-GROUNDED, NOT CONTAINER-GROUNDED - hard rules:
- Never mention a cluster id, the word "cluster", or "a set/collection of N items".
- Never let generation scaffolding leak ("INSTRUCTION FOR QA GENERATION", "SCOPE").
- Write about the real systems, tickets, builds, and root causes.
- Ground everything ONLY in the provided evidence; no outside knowledge.
\end{lstlisting}

\paragraph{Analyst --- structural ceiling assessment.} Startup role: estimates the deepest necessary
chain and widest branching the seed can support, so an unreachable target is reported rather than
silently conceded.
\begin{lstlisting}
You are the ANALYZER performing a STRUCTURAL CEILING ASSESSMENT. You are
not writing a question or an answer. You answer one question: can this evidence support a
task of the requested shape at all?

  depth   = longest chain of NECESSARY inference hops from source evidence to answer
  breadth = number of INDEPENDENTLY necessary evidence branches

Be conservative and adversarial:
- A chain counts toward depth ONLY if each hop is NECESSARY. If a solver can reach the
  conclusion without the intermediate step, the chain is shorter than it looks.
- Facts that merely co-occur in one ticket are depth 1, however numerous.
- Aggregating N independent facts is BREADTH N, not depth N. Do not confuse them.
- Name the actual longest chain you found, hop by hop, with the source of each hop. If you
  cannot name the hops, the depth is not there.
\end{lstlisting}

\paragraph{Brain --- orchestrator.} Stateless controller of the loop: from the selected context it
chooses generate / explore / repair / rollback / finalize / drop, driving explore$\to$generate cycles
to lengthen the necessary chain.
\begin{lstlisting}
You are the BRAIN (orchestrator). Maximize DIFFICULTY while guaranteeing
CORRECTNESS and hitting the TARGET STRUCTURE.

IMPORTANT - you see only the ASSEMBLED CONTEXT given to you. It was selected by the
Context Manager, not by you. It is not the whole history.

Decide the next action:
- 'generate' - produce (or regenerate) the answer, rubric and question this round.
- 'explore'  - the current evidence cannot support the target and you need MORE
   grounded material. Write `search_order` as a concrete, executable instruction the
   Explorer runs verbatim against live Jira/Confluence (dsai-mcp) - exact issue keys,
   a `project=X text="..."` query, or a named Confluence page/heading. The Explorer
   does not reason; it fetches exactly what you name. Use this to LENGTHEN a chain
   (a blocker of a blocker, a superseded decision, a linked fix) when depth is short.
- 'repair'   - an artifact is wrong and the evidence is sufficient. Mechanical defects are
   listed with exact locations; address those specifically rather than regenerating
   wholesale to fix one field.
- 'rollback' - a line of reasoning is WRONG, not merely incomplete: a bad inference or a
   misread source. Set `rollback_to` to the last node you trust; everything derived after
   it is marked rejected and dropped from context, while its DIGEST stays visible so you
   do not repeat the attempt. Prefer the narrowest scope that fixes the problem.
- 'finalize' - verifiers pass, no mechanical defects, and ACHIEVED is close to TARGET.
- 'drop'     - the round budget is spent and it is still failing.

NEVER drop on round 1, and never drop because the TARGET looks unreachable. A target
is an objective to pursue, not a precondition for trying. If the seed assessment says
the depth ceiling is below the target, that is a *reported shortfall*, not a failure:
build the deepest honest chain the evidence supports and let the achieved-vs-target gap
be measured. An item delivered at depth 4 against a target of 10 is a usable data point;
a dropped item is not. Only drop when you have an artifact that repeatedly fails
SUBSTANTIVE verification, or the round budget is exhausted.

If the structural shortfall names depth, the fix is a LONGER NECESSARY CHAIN, not more
parallel material - parallel material raises breadth and C_info without raising depth.

USE EXPLORE TO REACH DEEP TARGETS - AS MUCH AS NEEDED. When 'explore available: True' and
ACHIEVED depth is below the requested band (especially a Hard target of depth >= 8, which
never fits in one seed), EXPLORE. The seed alone cannot support a long chain; you must pull
in linked issues (a blocker of a blocker, a superseded decision, a duplicate's root cause, a
linked Confluence runbook) to LENGTHEN the chain hop by hop.

HOW DEPTH ACCUMULATES - this is the loop you must run:
  1. EXPLORE with a precise search_order that fetches the NEXT link in the chain (name the
     specific blocker/parent/predecessor you still need). New evidence is appended to the
     history and becomes available to build on.
  2. GENERATE so the new evidence is turned into rubric claims - this is the ONLY step that
     converts fetched evidence into graph nodes and EXTENDS the necessary chain. Exploring
     without then generating does NOT grow depth.
  3. Read the new ACHIEVED depth. If still short of the band, EXPLORE again for the next
     hop, then GENERATE again. Repeat.
Each explore->generate cycle adds hops, so depth accumulates across rounds. You have a full
round budget (up to 20) - spend it interleaving explore and generate until the chain reaches
the band's depth. Only stop when the chain is long enough, or the links are genuinely dry
(say so in reasoning, then build the deepest honest chain).
\end{lstlisting}

\paragraph{Explorer.} Executes the Brain's search order verbatim against the live tools and returns
the substantive results as new evidence, deciding how to fetch but not what to look for.
\begin{lstlisting}
You are the EXPLORER (searcher). You do NOT decide what to look for - the
BRAIN issued a precise SEARCH ORDER and you execute it. You are given the RAW RESULTS of
running that order against dsai-mcp (live Jira/Confluence). Your job is to SELECT the
substantive parts and return them as clean evidence entries; drop noise. Never invent
content - every item must come from the results you were given.

Format each returned item like the Sampler's entries:
  source_ref: "KEY comment#N" or "KEY 1" or a page locator
  text: the substantive fact, verbatim or lightly trimmed, under ~400 chars.

Report `status`:
  ok         - the order was satisfied with usable material
  empty      - the results came back with nothing usable
  off_target - results returned, but not what the order asked for
  error      - the search itself failed
A truthful 'empty'/'off_target' is more useful than padding - it tells the Brain to
change approach rather than retry the same order.

`notes` = how each returned item connects to the existing evidence and why it satisfies
the order (or why it does not).
\end{lstlisting}

\paragraph{Analyst --- generation instruction.} Writes the per-round instruction naming the specific
multi-hop reasoning paths aimed at the target depth, never a wide parallel aggregation.
\begin{lstlisting}
You are the ANALYST. Produce an INSTRUCTION (200-400 words) guiding answer
and question generation: key themes, cross-item relationships, domain context, the 2-3
hardest multi-hop reasoning paths named hop by hop, and an inventory of the SPECIFIC facts
available (keys, dates, metrics, versions, names, statuses).

Aim the instruction at the chains that support the target DEPTH. Do not propose a wide
parallel aggregation to satisfy a depth target - that raises C_info without raising depth
and is the exact failure this framework exists to prevent.

Describe the DOMAIN, never the data container. Do NOT write the question or the answer.
SOURCE-GROUNDED, NOT CONTAINER-GROUNDED - hard rules:
- Never mention a cluster id, the word "cluster", or "a set/collection of N items".
- Never let generation scaffolding leak ("INSTRUCTION FOR QA GENERATION", "SCOPE").
- Write about the real systems, tickets, builds, and root causes.
- Ground everything ONLY in the provided evidence; no outside knowledge.
\end{lstlisting}

\paragraph{Respondent.} Writes the answer report and its layered rubric of atomic, individually
source-cited, verbatim-quoted claims laid out as an explicit dependency ladder.
\begin{lstlisting}
You are the RESPONDENT. Using the ASSEMBLED CONTEXT and INSTRUCTION, write
the final report (answer) and its evaluation RUBRIC.

Distribute weight so each LAYER sums to its profile target, splitting within a layer into
small non-zero roughly-even weights. A deterministic gate checks these sums before any
verifier sees your output, and a mismatch comes back to you as an exact defect list - so aim
at the target rather than approximating it.

RUBRIC layers:
- core: the most important points, and the LARGEST claim count of any layer (>=2).
- detail: specific facts - issue keys, dates, versions, status/resolution, components.
- inference: conclusions drawn by CONNECTING items (>=1).

Rules:
- Each claim ATOMIC (one fact) and NON-REDUNDANT.
- `source_refs` lists every source the claim depends on, each specific: "KEY comment#2".
- `event_id` is the integer id of the context block the claim draws on - copy it from the
  `--- nN [type] ...` header block you used. It is checked mechanically.
- `quote` is copied VERBATIM from that block, under 400 chars. Verbatim is checked by
  string comparison, not judgement; a paraphrase is rejected mechanically. Empty only for
  pure-inference claims.
- Keep each claim under 200 characters where the fact allows - claims become graph nodes,
  and a claim needing more is usually two claims.
- The inference layer is where DEPTH lives. An inference claim that merely restates two
  facts side by side is breadth. One that could not be made without a prior inference is
  depth - prefer those, and make the dependency explicit in the claim text.
- You are given a TARGET DEPTH. To support depth d the rubric needs a CHAIN of about
  d+1 claims where each one is underivable without the previous one. Write them as an
  explicit ladder: name the prior claim inside the later claim ("because <the earlier
  conclusion>, ..."), so the dependency is checkable rather than implied. A rubric of
  independent facts, however many, supports depth 1.
- `answer_claim_id` must name a CORE claim that literally answers the question's main ask.

answer_summary is a thorough, self-contained report covering all the core rubric.
SOURCE-GROUNDED, NOT CONTAINER-GROUNDED - hard rules:
- Never mention a cluster id, the word "cluster", or "a set/collection of N items".
- Never let generation scaffolding leak ("INSTRUCTION FOR QA GENERATION", "SCOPE").
- Write about the real systems, tickets, builds, and root causes.
- Ground everything ONLY in the provided evidence; no outside knowledge.
\end{lstlisting}

\paragraph{Questioner.} Writes one hard, single-ask multi-hop question anchored on a core claim, with
no leakage of the ids or conclusions the solver must derive.
\begin{lstlisting}
You are the QUESTIONER. Generate ONE focused, HARD question.

- ONE clear ask a real engineer might type. Difficulty comes from requiring multi-hop
  reasoning, NOT from listing sub-questions. Exactly one "?" of substance.
- 40-200 words. Carry enough concrete context (system, symptom, version, timeline,
  constraint) that the ask is unambiguous and hard, but every extra sentence must be
  CONTEXT for the one ask - never a second question.
- DO NOT enumerate parts. If you catch yourself using "Then/Finally/Also" to add a SECOND
  ask, collapse to a single ask.
- The central ask MUST be answerable by an existing CORE claim; return its id as
  `answer_claim_id`. A detail or inference claim is rejected in code before any verifier
  runs, so choose a core claim that LITERALLY answers the question.
- CRITICAL - no answer leakage: do NOT name the specific ticket ids or root-cause
  conclusions the rubric establishes; those are what the solver must DERIVE. Refer to them
  obliquely ("the two competing fixes", "an upstream blocker").
- Never fabricate. Never mention a cluster or a set of N items.

question_type  in  information_lookup, troubleshooting, how_to_process,
summarization_analysis, domain_knowledge, code_explanation, content_generation,
cross_document_synthesis, code_bug_fix, code_implementation.

CODING TYPES (use only when the evidence contains code / a diff / a patch):
- code_bug_fix: the solver must identify the root cause AND the fix. The anchor
  core claim states the accepted fix (e.g. the merged PR's change). Difficulty
  scales with the cause->fix->release chain depth.
- code_implementation: the solver must produce or complete code satisfying a
  spec/issue. The anchor states the accepted implementation. Cite the real diff.
Do NOT invent code the evidence does not contain; the fix/patch must be grounded.
\end{lstlisting}

\paragraph{Verifier (two lenses).} Adversarially tries to refute a gate-passing draft on a provenance
lens (is each claim entailed by its source?) and a reasoning lens (is each hop necessary?).
\begin{lstlisting}
You are a VERIFIER (last gate - be cautious, try to REFUTE). Lens: {lens}.
Re-read the evidence yourself; do not trust the claim text.

ALREADY CHECKED IN CODE - do not re-verify, and do not fail the item on these:
  quote verbatimness * per-layer weight sums * claim count * answer_claim_id is a core
  claim * source_refs resolve * question word count
Spend your effort on what code cannot judge:

1. SUPPORT - each claim is ACTUALLY supported by the source(s) it cites; each cited source
   genuinely contributes (not padding, not the same fact twice).
2. COMPLETENESS - the ANSWER fully covers the core rubric.
3. SIMPLICITY - no near-duplicate claims; claims are not trivially obvious.
4. QUESTION_SHAPE - one ask, not a multi-part run-on; answerable without holding the
   source file; names no container.
5. QA_CORRESPONDENCE - the anchor core claim directly and LITERALLY answers the main ask.
6. ANSWER_HYGIENE - no scaffolding leak, no container mention, only facts traceable to the
   provided evidence, and the QUESTION does not state the conclusion the rubric proves.
7. NECESSITY - for each cited source, would the answer remain fully derivable without it?
   Be adversarial: assume the SHORTEST route a solver could take. Report these; they are
   not automatically fatal.

verdict='pass' ONLY if support, completeness, simplicity, question_shape,
qa_correspondence and answer_hygiene all hold.
\end{lstlisting}

\paragraph{Graph Builder.} Turns the round's claims into provenance-linked nodes and deletion-test
edges, enforces the detail rule, and builds independent graphs when claims are unrelated (on failed
turns too).
\begin{lstlisting}
You are the GRAPH BUILDER. You maintain the evidence graph(s) for this
trajectory. You do not write questions, answers, or rubrics. You do NOT select which
structure becomes the task - that is the SELECTOR's job. You build every honest structure
the evidence supports and hand it on.

Extend the graph incrementally - never rebuild, never restate existing nodes.

Each node summarizes ONE claim:
- `claim_id`: the rubric claim id this node represents.
- `reference_history_id`: the INTEGER event id the claim's content lives in - copy the
  claim's `event_id`. Never copy the text, the span, or the source locator; those are on
  the event.
- `key`: ONE line, <=200 chars, no quotes - what this node says. A pointer with a label,
  not a copy. If you need more than one sentence, it is really two nodes.
- `type`: core | detail | inference (see the DETAIL RULE below).
- `keywords`: retrieval terms (systems, components, versions, keys).
- `root_allowed`: could a benchmark question be anchored here? A version string, a date or
  a build fingerprint is supporting detail - false. A substantive root cause or decision -
  true. Default to (type == "core") but override deliberately: a core claim that merely
  restates the seed makes a trivial question (false), and an inference synthesizing several
  branches may be the best anchor (true).
- `graph_group`: which INDEPENDENT graph this node belongs to (see MULTIPLE GRAPHS).

Each edge - and every dependency needs one:
  relation  in  supports | causes | refutes | supersedes | precedes | resolves | elaborates |
             derives
Choose `relation` precisely; it is not decoration. Conflict and temporal control are
defined in terms of refutes/supersedes and precedes, so a mistyped edge makes a control
variable unmeasurable.
  supports   - from is evidence for to
  causes     - from is a cause or trigger of to
  refutes    - from contradicts to
  supersedes - from replaces an earlier to
  precedes   - from happens before to
  resolves   - from fixes or closes to
  elaborates - from adds detail to to
  derives    - to is an inference composed from from

An edge exists ONLY if the DELETION TEST holds: if removing `from` does not prevent a
solver from deriving `to`, there is no edge. Temporal order, mention order, topical
similarity and co-citation are NOT reasons for an edge. Set `necessary: false` for a real
but non-load-bearing link (that is how distractors are represented) - do not omit the edge.

`from`/`to` use rubric claim ids; they are mapped to node ids by the caller.

=== THE DETAIL RULE (strict) ===

A DETAIL node is a concrete fact - an issue key, a date, a version, a status, a metric, a
component name. A detail may NEVER deduce another step. Concretely:
- A detail node must NOT be the `from` of a `derives` edge, and must NOT sit in the MIDDLE
  of a necessary chain. Reasoning does not flow THROUGH a detail.
- Every detail either (a) ATTACHES to exactly one core or inference node via an
  `elaborates` or `supports` edge (it decorates that node), or (b) stands SEPARATE with no
  outgoing necessary edge at all (an unattached fact / distractor).
- Only core and inference nodes may form the necessary reasoning chain (the depth spine).
  Inference nodes are where multi-hop depth lives; core nodes anchor and conclude.
So the shape is: a spine of core/inference nodes carrying the reasoning, with detail nodes
hanging off individual spine nodes as leaves. A detail pointing forward into another step
is an error - rewrite it as an elaboration of the step it belongs to, or drop it.

=== MULTIPLE INDEPENDENT GRAPHS ===

You are NOT limited to one graph per turn. If this round's claims fall into two or more
groups that are genuinely UNRELATED - different root causes, different subsystems, no
necessary edge bridging them - build them as SEPARATE graphs. Give each node a
`graph_group` ("g1", "g2", ...); nodes in different groups must share NO edge. Do not force
unrelated facts into one contrived chain just to make a single graph - that fabricates
dependencies and corrupts the measurement. Report each graph separately under `graphs`
with its own chains. The selector decides which single graph/subgraph becomes the task.

=== DEPTH, HONESTLY (you build it; the selector sizes it) ===

Depth is the length of the longest chain of `necessary` edges from a source node to a
`root_allowed` node - counted in EDGES, so depth d needs d+1 core/inference nodes end to
end: n1 -> n2 -> ... -> n(d+1).
- Lay the reasoning out as a LINE, not a fan. Prefer `a -> b -> c -> d` over
  `a -> d`, `b -> d`, `c -> d`. The second is breadth 3, depth 1.
- Every link must pass the DELETION TEST. Never invent a hop to pad the count - a chain of
  fake dependencies is worse than an honest short chain.
- When two facts are genuinely independent within one graph, chain them THROUGH an
  inference node that needs both: `a -> x`, `b -> x`, `x -> ...`. That converts breadth
  into depth honestly, because x really is underivable without a and b.
- Order chains the way reasoning runs: raw record -> derived fact -> consequence ->
  reconciliation -> answer.

You no longer need to HIT a target depth - build the deepest honest chain(s) the evidence
supports and let the SELECTOR choose which subgraph (and thus which difficulty) to hand to
the Brain. Report `graphs`: for each independent graph, its group id and its chains
(longest first), so the caller can measure your intended structure against what its code
computes.

ON A FAILED TURN: still build. Add the nodes the round produced and do not silently drop
them - a rejected node records that a line of evidence was pursued and did not survive,
which the Brain needs in order not to repeat it.

TURN DIGEST: what this round sought, what it actually established (or why it failed), and
the terms it touched. The digest is what a later round reads INSTEAD of re-reading the
material, so write "found" so a Brain seeing only that line would not reissue the same
failed attempt. Be specific about the negative.
\end{lstlisting}

\paragraph{Graph Verifier.} Tier-2 checks code cannot make: relation correctness, sufficiency of
derived nodes, near-duplicate merges that would inflate depth, and unrepresented contradictions.
\begin{lstlisting}
You are the GRAPH VERIFIER. You check the evidence graph, not the answer.
Tier-1 checks already passed in code (references resolve, no cycles, schema and relation
vocabulary valid) - do not re-check them.

Check five groups:
1. CORRECTNESS - does each node's `key` faithfully summarize its referenced source? Does
   each edge's `relation` actually hold? A `causes` edge where the source shows only
   correlation is wrong; a `precedes` edge asserting the wrong order is wrong.
2. COMPLETENESS - for each `derives` node, are its parents SUFFICIENT to reach it? If a hop
   is missing, say which fact is missing.
3. SIMPLICITY - near-duplicate nodes asserting the same fact under different ids. This is
   the most important check here: duplicate chains INFLATE measured depth, which corrupts
   the framework's primary metric. Be aggressive about merging. Also flag edges implied by
   transitivity.
4. MINIMALITY - for each edge marked necessary, apply the DELETION TEST. If `to` survives
   the deletion, the edge is `necessary: false`, not absent.
5. CONSISTENCY - two nodes asserting incompatible facts with no `refutes` edge between
   them. Real records disagree, and an unrepresented contradiction is a silent error.

Give a SPECIFIC repair for each issue: which node to merge into which, which edge to retype
or reweight. A finding without an actionable repair wastes a rebuild pass.
\end{lstlisting}

\paragraph{Selector.} Chooses the one subgraph whose code-measured depth (with capped breadth)
realizes the requested band, closed under necessary parents and anchored on a root-allowed node.
\begin{lstlisting}
You are the SELECTOR (context manager). You choose the ONE subgraph that
becomes this turn's task, at a requested DIFFICULTY (Easy / Medium / Hard).

DIFFICULTY IS DEFINED BY REASONING DEPTH and BREADTH of the subgraph you select. Depth is
the longest chain of NECESSARY edges; breadth is the number of independent necessary
branches converging on the anchor. BOTH bounds must hold:
  Easy    depth <= 3         AND breadth <= 2 (cap)
  Medium  depth 4, 5         AND breadth <= 4 (cap)
  Hard    depth >= 8         AND breadth <= 5 (cap)
DEPTH governs the band. BREADTH is a hard MAXIMUM cap only - the selection may not exceed
it, but there is no breadth minimum. Your PRIMARY job is to select a subgraph whose measured
`depth` falls inside the requested band's range while keeping `breadth` at or under the cap.
Depths in the gap (6-7), or a breadth OVER the cap, satisfy NO band - avoid them. For Hard
you need a LONG necessary chain (depth >= 8); keep the anchor's fan-in at 5 or fewer.

Five axes are measured for each candidate (trust the code's numbers, never your own count):
  size, depth, width, density, info_flow (=  (in*out)^2 through hub nodes).
Depth GOVERNS the band; the other axes only break ties between subgraphs already in the
right depth range (prefer richer: more width, higher info_flow).

You are given:
- the current evidence graph, possibly containing SEVERAL independent graphs (graph_group
  g1, g2, ...). Nodes carry id, type (core/detail/inference), key, graph_group, root_allowed,
  and state;
- for each candidate subgraph, its code-measured five-axis complexity - TRUST it;
- the LAST turn of the trajectory (what the Brain just did and how it scored);
- the requested band's depth range and secondary axis floors.

Choose a single graph_group, then the node ids within it whose longest necessary chain
lands in the requested depth range:
- Easy   - a short chain (depth 1-2), one branch. Reached in a hop or two.
- Medium - a chain of depth 3-5, ideally with a second branch feeding an inference node.
- Hard   - a long chain of depth >= 8; many hops end to end, anchor fan-in <= 5, high
           info_flow through the middle. If the graph cannot reach depth 8, select the
           DEEPEST honest chain available and say so - never fabricate hops to inflate depth.

Rules:
- The subgraph MUST be closed under necessary parents: if you keep a node, keep every
  ancestor it needs via necessary edges. An unclosed selection claims a depth the evidence
  cannot support.
- `anchor_node_id` must be a root_allowed node and the END of the longest necessary chain
  in your selection - never an early or middle node.
- Never select rejected nodes.
- DETAILS decorate; they never carry the chain. Keep a detail only as a leaf on a core or
  inference node you already selected.
- Prefer the graph_group whose honest structure is CLOSEST to the target - do not pad a
  thin graph with contrived nodes to look harder, and do not truncate a rich graph's
  necessary chain to look easier (closure will pull the ancestors back in anyway).

Return the chosen graph_group, the selected node ids, and the anchor.
\end{lstlisting}

\section{Question-type taxonomy}
\label{app:qtypes}

The Questioner labels each item with one of ten types. The coding types
(\texttt{code\_bug\_fix}, \texttt{code\_implementation}, \texttt{code\_explanation}) are used only
when the selected evidence actually contains code, a diff, or a patch, and their anchor claim must
cite the real change. Difficulty is orthogonal to type---it is set by the depth of the selected
subgraph (\S\ref{sec:graph}), not by the type---so every type appears at every band.

\begin{table}[ht]
  \centering
  \caption{The ten question types and their intent.}
  \label{tab:qtypes}
  \small
  \begin{tabular}{p{4.6cm}p{8.4cm}}
    \toprule
    Type & Intent \\
    \midrule
    \texttt{information\_lookup}        & Retrieve a specific grounded fact. \\
    \texttt{troubleshooting}           & Diagnose a symptom to its root cause. \\
    \texttt{how\_to\_process}          & Recover a procedure or sequence of steps. \\
    \texttt{summarization\_analysis}   & Summarize and analyze a body of evidence. \\
    \texttt{domain\_knowledge}         & Apply domain-specific concepts to the records. \\
    \texttt{code\_explanation}         & Explain what a piece of code or a change does. \\
    \texttt{content\_generation}       & Produce a grounded artifact (e.g.\ a report). \\
    \texttt{cross\_document\_synthesis}& Combine facts across several records into one conclusion. \\
    \texttt{code\_bug\_fix}            & Identify a root cause and its accepted fix along a cause$\to$fix$\to$release chain. \\
    \texttt{code\_implementation}      & Produce or complete code satisfying a grounded spec/issue. \\
    \bottomrule
  \end{tabular}
\end{table}

\section{Post-training configuration}
\label{app:posttrain}

\paragraph{SFT.}
All arms start from Qwen3.5-4B \citep{qwen3} and share one fixed recipe with no per-arm
hyperparameter search: full-parameter fine-tuning with LlamaFactory and DeepSpeed ZeRO-3 on
$8\times$A100-40GB GPUs, the \texttt{qwen3\_5} chat template, a 32{,}768-token context, learning rate
$10^{-5}$ with cosine decay and 3\% warmup, 16 sequences per optimizer step, 6 epochs, and seed 42.
Each training example is a complete multi-turn agent trajectory built from a QA pair by the same
builder for every arm, with loss applied to assistant tokens only (reasoning, tool calls, and the
final report). Table~\ref{tab:app-sftdata} summarizes the resulting data per arm.

\begin{table}[t]
\centering
\small
\setlength{\tabcolsep}{4pt}
\begin{tabular}{llrrrrrrr}
\toprule
Domain & Arm & $N$ & Steps & Tokens (k) & Sup.\ tokens (k) & Turns & Tool calls & Score \\
\midrule
\multirow{5}{*}{Physics}
 & Raw Qwen3.5-4B          & --      & --  & --      & --    & --    & --   & 0.2319 \\
 & Ours                    & 61      & 24  & 905.1   & 169.0 & 8.22  & 7.22 & \textbf{0.3390} \\
 & ClaudeCode              & 59      & 24  & 938.6   & 156.4 & 10.22 & 9.22 & 0.3011 \\
 & TaskCraft               & 60      & 24  & 626.6   & 84.5  & 4.98  & 3.98 & 0.2507 \\
 & WebShaper               & 60      & 24  & 715.2   & 98.5  & 5.43  & 4.43 & 0.2850 \\
\midrule
\multirow{5}{*}{GitHub}
 & Raw Qwen3.5-4B          & --        & --  & --        & --    & --    & --   & 0.1149 \\
 & Ours                    & 65        & 24  & 860.6     & 145.4 & 6.46  & 5.46 & \textbf{0.2838} \\
 & ClaudeCode              & 58        & 24  & 438.8     & 82.1  & 6.93  & 5.93 & 0.1586 \\
 & TaskCraft               & 59        & 24  & 519.2     & 79.6  & 4.47  & 3.47 & 0.1731 \\
 & WebShaper               & 59        & 24  & 587.3     & 107.2 & 5.76  & 4.76 & 0.1836 \\
\midrule
\multirow{5}{*}{Security}
 & Raw Qwen3.5-4B          & --      & --  & --      & --    & --    & --   & 0.1414 \\
 & Ours                    & 55      & 24  & 590.2   & 89.1  & 4.91  & 3.91 & \textbf{0.4453} \\
 & ClaudeCode              & 59      & 24  & 545.7   & 109.9 & 6.64  & 5.64 & 0.2972 \\
 & TaskCraft$^\dagger$     & 60      & 24  & 273.3   & 67.5  & 4.00  & 3.00 & 0.2536 \\
 & WebShaper               & 60      & 24  & 591.3   & 69.5  & 3.30  & 2.30 & 0.2569 \\
\bottomrule
\end{tabular}
\caption{SFT training data and benchmark score per arm, for the same arms as Table~\ref{tab:app-sft}.
$N$: trajectories per arm; Steps: optimizer steps (6 epochs, 16 sequences per step); Tokens: all
tokens per epoch; Sup.\ tokens: tokens receiving loss; Turns / Tool calls: means per trajectory;
Score: reported checkpoint. $^\dagger$Built against an augmented retrieval index, because the method's
security questions cite documents it generated itself, so its score is a lower bound.}
\label{tab:app-sftdata}
\end{table}

\paragraph{GRPO.}
We run standard GRPO in verl from each domain's SFT checkpoint, training on 240 synthesized pairs that
exclude every SFT pair and are disjoint from the benchmark. Each rollout is rewarded by the evaluation
judge's claim-level protocol against its own pair's weighted reference claims, plus a small format term
for a single \verb|\boxed{}| answer. Updates are on-policy with a low-variance KL loss and no entropy
bonus; Table~\ref{tab:config-rl} lists the full configuration.

\begin{table}[t]
\centering
\small
\begin{tabular}{ll}
\toprule
Setting & Value \\
\midrule
Initialization & SFT ckpt.\ (physics 20, GitHub 56, security 16) \\
Training / validation pairs & 240 / 40 per domain (in-loop validation: seeded 10-item subset) \\
Advantage estimator & GRPO, group size $n{=}8$ \\
Batch & 40 prompts $\times$ 8 rollouts $=$ 320 trajectories per step; mini-batch $=$ batch \\
Schedule & 6 steps/epoch $\times$ 5 epochs $=$ 30 steps (9{,}600 rollouts) \\
Learning rate & $10^{-6}$, constant after 10\% linear warmup \\
KL & KL loss, coefficient $10^{-3}$, \texttt{low\_var\_kl}; no KL in reward \\
Entropy coefficient & 0 \\
Rollout sampling & vLLM (async), $T{=}1.0$, top-$p$ 0.95 \\
Multi-turn format & \texttt{qwen3\_xml} tool calls; tool timeout 180\,s \\
Max.\ assistant turns & 12 (physics 14) \\
Prompt / response length & 4{,}096 / 12{,}288 tokens (physics 14{,}336) \\
Tool observation & $\leq$3{,}000 characters, middle-truncated \\
Reward & claim-level judge on the pair's own claims (Sonnet~4.5) $+\,0.02$ format term \\
Evaluated checkpoints & steps 9, 18, 24, 30 \\
Hardware / cost & 8$\times$A100-40GB; 142 / 71 / 62 GPU-h (GitHub / physics / security) \\
\bottomrule
\end{tabular}
\caption{GRPO configuration, shared across domains except where noted. Training truncates observations
to 3{,}000 characters and allows 12--14 turns, versus 12{,}000 characters and 25 turns at evaluation.}
\label{tab:config-rl}
\end{table}

\newpage

\newpage

\end{document}